\documentclass{INTERSPEECH2023}

 \interspeechcameraready 

\title{Cross-lingual Knowledge Transfer and Iterative Pseudo-labeling for Low-Resource Speech Recognition with Transducers}
\name{Jan~Silovsky, Liuhui~Deng, Arturo~Argueta, Tresi~Arvizo, Roger~Hsiao, Sasha~Kuznietsov, Yiu-Chang~Lin, Xiaoqiang~Xiao, Yuanyuan~Zhang}
\address{
  Apple
}
\email{jsilovsky@apple.com}

\begin{document}

\maketitle
 
\begin{abstract}

Voice technology has become ubiquitous recently. However, the accuracy, and hence experience, in different languages varies significantly, which makes the technology not equally inclusive. The availability of data for different languages is one of the key factors affecting accuracy, especially in training of all-neural end-to-end automatic speech recognition systems.

Cross-lingual knowledge transfer and iterative pseudo-labeling are two techniques that have been shown to be successful for improving the accuracy of ASR systems, in particular for low-resource languages, like Ukrainian. 

Our goal is to train an all-neural Transducer-based ASR system to replace a DNN-HMM hybrid system with no manually annotated training data. We show that the Transducer system trained using transcripts produced by the hybrid system achieves 18\% reduction in terms of word error rate. However, using a combination of cross-lingual knowledge transfer from related languages and iterative pseudo-labeling, we are able to achieve 35\% reduction of the error rate.

\end{abstract}

\section{Introduction}

It has been decades since speech technology reached the maturity allowing creation of user facing products. At the same time, ASR, like other machine learning fields, has seen tremendous progress in the past decade. Especially with the advent of deep neural networks (DNNs).

Conventionally, the ASR systems consist of an acoustic model and a language model. Acoustic models are typically based on Deep Neural Network - Hidden Markov Model (DNN-HMM) hybrid architecture~\cite{dnn_hmm} and the language models are represented by n-gram model.

In this paper, we investigate development of an end-to-end (E2E) ASR system based on Transformer Transducer~\cite{DBLP:journals/corr/abs-1211-3711} architecture for Ukrainian.

The accuracy of Automatic Speech Recognition (ASR) systems is usually highly dependent on the availability of high-quality annotated training data.
In many low-resource languages, such as Ukrainian, the amount of annotated data is limited, which makes it challenging to train accurate ASR models.
One way to tackle the lack of annotated data is to use unsupervised pre-training~\cite{unsupervised_pretraining}, which has earned a lot of attention in the past years.
In speech recognition, these techniques are represented for example by wav2vec~\cite{wav2vec}, followed by wav2vec 2.0~\cite{wav2vec2}.
However, for many languages, including Ukrainian, it is not just the amount of annotated data that is limited, but the availability of data in general, which is significantly lower than for languages like English, Mandarin, or even Russian.

We aim to minimize the amount of data that needs to be annotated by humans, and hence, we assume availability of no manually annotated data.
On the other hand, we assume availability of automatic transcripts produced by a DNN-HMM system, which is to be replaced by the Transduced-based system.

To cope with the limited amount of data, we investigate cross-lingual knowledge transfer. While the unsupervised pre-training can be also applied in a cross-lingual fashion~\cite{crosslingual_wav2vec2}, the results presented in~\cite{xu2020iterative} suggest that pseudo-labeling leads to better accuracy than unsupervised pre-training and hence we investigate a combination of cross-lingual semi-supervised learning and iterative pseudo-labeling.

Our reasoning is that semi-supervised pre-training an ASR model on a large amount of data from related languages can help improve the performance of the model when fine-tuned on a more constrained amount of data from the target language.

Additionally, we use iterative pseudo-labeling to further improve the performance of the model by using the model's own predictions to generate transcripts with better accuracy for the next iteration of the training.

\section{Related work}

\subsection{Cross-lingual knowledge transfer}

In the context of ASR, cross-lingual knowledge transfer aims to bootstrap or improve models for low-resource languages by using multiple rich-resource languages~\cite{ghoshal2013multilingual,huang2013cross,cho2018multilingual,pratap2020massively}. The idea is that the model learns to extract general acoustic and linguistic features that can be useful across languages.
These language-agnostic features can then be fine-tuned on data from the target language to improve performance on that language. 

In~\cite{ghoshal2013multilingual}, authors deal with training of a multilingual DNN-HMM. They propose a training curriculum in which the model is gradually fine-tuned with data from multiple languages, one language at a time, and eventually fine-tuned to the target language.

Alternatively, the authors in~\cite{huang2013cross,heigold2013multilingual} propose training a multilingual DNN models on a pool of multilingual training data from scratch, sharing all the layers except for the language-specific output layers.
The reported results indicate that features learned by hidden layers are transferable across languages and lead to better accuracy compared to monolingual baselines. 

In~\cite{cho2018multilingual}, the authors present multi-lingual pre-training of a prior sequence-to-sequence CTC model with an attention decoder using a set of 10 languages, and then porting the model to another four target languages using transfer learning. The authors propose multi-stage transfer learning. First, fine-tuning only the decoders of the prior model to the target languages, and then fine-tuning the whole model, including the encoder, in the subsequent stage.

In~\cite{pratap2020massively}, the authors train a massive E2E multilingual model with 1 billion parameters on 50 languages. When compared to monolingual models, a degradation is observed on high-resource languages, while an improvement is observed on low-resource languages. The prior model is also fine-tuned to unseen low-resource languages, outperforming monolingual baselines.

\subsection{Pseudo-labeling}

Pseudo-labeling is a semi-supervised learning~\cite{chapelle2009semi} approach that uses the output of a pre-trained model to create labeled data that can be used to train a more accurate model. This approach has been shown to be effective in ASR applications, where labeled data are often limited and expensive to obtain~\cite{synnaeve2019end,kahn2020self}.

The quality of pseudo-labels also plays a crucial role, where confidence-based data selection selects queries with higher scores, which leads to better model accuracy~\cite{synnaeve2019end,yu2010unsupervised,huang2013semi}.

In recent years, there has been increased interest in iterative pseudo-labeling (IPL)~\cite{chen2020semi,xu2020iterative,likhomanenko2020slimipl}.
The motivation behind IPL is an assumption that the model from one iteration can be used to re-decode data for the next iteration of the training with higher accuracy, and hence improve the model trained in the next update.
One key consideration in IPL is the frequency of the model updates.
Higher frequency can avoid re-decoding with a stale model and result in re-decoding less data overall, but on the other hand can lead to increased GPU computation required for the model updates.
The decision is typically based on a number of factors including the accuracy of the prior model and expected accuracy gains between the model updates.

\section{Experimental setup}

\subsection{Model architecture}

Our target model~\cite{pawel_variable_attention_TT,thien_bilingual_transducer} is a Transducer~\cite{DBLP:journals/corr/abs-1211-3711} with a Conformer~\cite{https://doi.org/10.48550/arxiv.2005.08100} acoustic encoder and Transformer~\cite{https://doi.org/10.48550/arxiv.1910.12977,https://doi.org/10.48550/arxiv.2002.02562} label encoder. The model also uses a convolutional downsampling front-end. In total, the model has approximately 100M parameters.

\subsection{Data}

In this paper, we use the term semi-supervised data for data which are not manually transcribed but accompanied with transcripts produced by an ASR system, possibly previously trained on supervised data.

To explore cross-lingual knowledge transfer, we used anonymized and randomly sampled semi-supervised data from 5 Slavic languages --- Czech, Polish, Russian, Slovak and Ukrainian. We built two datasets, a) naturally weighted (NW) and b) balanced (BL). Table~\ref{tab:semisup_data} shows the amount of data in the datasets. The Naturally weighted dataset reflects the amount of available data with respect to their natural frequency and is heavily unbalanced. Thus, we constructed the balanced dataset which aims at uniform representation of the languages. 
The same period is covered for all languages.
Data for more frequent languages are randomly sub-sampled.

\begin{table}[th]
  \caption{Amounts of semi-supervised data in Naturally weighted and Balanced datasets}
  \label{tab:semisup_data}
  \centering
  \begin{tabular}{ l r r }
    \toprule
    & Naturally weighted (NW) & Balanced (BL)\\
    &  [khrs] & [khrs]\\
    \midrule
    UKR & 12 & 25 \\
    RUS & 501 & 16 \\
    POL & 92 & 27 \\
    CZE & 35 & 23 \\
    SVK & 8 & 18 \\
    \midrule
    Total & 636 & 109 \\
    \bottomrule
  \end{tabular}
\end{table}

\subsection{Word-piece tokenizer}

Table~\ref{tab:tokenizers} lists three different sub-word tokenizers used in our experiments. All tokenizers were built using Sentencepiece~\cite{https://doi.org/10.48550/arxiv.1808.06226}, employing the byte-pair-encoding (BPE) algorithm~\cite{sennrich-etal-2016-neural}.
To avoid explosion of output dimensionality for the multi-lingual models and exploit the fact that the languages belong to the same family of Slavic languages, we do not combine mono-lingual tokenizers to build the multi-lingual tokenizers but we build a tokenizer from pooled transcripts, using the balanced dataset.

\begin{table}[th]
  \caption{Tokenizers}
  \label{tab:tokenizers}
  \centering
  \begin{tabular}{ l r r r }
    \toprule
    Tokenizer & Languages & Characters & Size \\
    \midrule
    5 Langs & CZE, POL, RUS & Latin, Cyrillic & 8k \\
     & SVK, UKR &  &  \\
    2 Langs & RUS, UKR & Cyrillic & 6k \\
    UKR & UKR & Cyrillic & 6k \\
    \bottomrule
  \end{tabular}
\end{table}

\section{Results}

We present our results in terms of relative Word Error Rate Reduction (WERR) where the reference is represented by the the DNN-HMM hybrid systems.
We note that the DNN-HMM system operates in streamable fashion but for the Transducer, results reported in this work are for a model configuration using global attention and hence not suitable for streaming. Hence, the reported WERR figures are not meant to contrast the accuracy of the DNN-HMM and Transducer-based E2E systems but to assess the impact of cross-lingual knowledge transfer and iterative pseudo-labeling for the latter systems.

Our training is a multi-stage process, where models trained in one stage are used to initialize models' parameters in the next stage.
We assign each model a reference identifier (model ref) for easier differentiation of the models.

\subsection{Baseline}

\begin{table}[th]
  \caption{Comparison of Ukrainian models trained with multi-lingual and mono-lingual tokenizers.}
  \label{tab:tokenizers_comparison}
  \centering
  \begin{tabular}{ l r r }
    \toprule
Tokenizer &	5 Langs & UKR \\
    \midrule
WERR [\%] & 17.4 & \textbf{18.1} \\
    \bottomrule
  \end{tabular}
\end{table}

We start with establishing reference results by training a mono-lingual Transducer model using the semi-supervised Ukrainian data only.

We hypothesize that the cross-lingual training, on which we report later, has an advantage of more data being available, but since we do not simply combine mono-lingual tokenizers to build the multi-lingual tokenizers in order to not explode the output dimensionality, our multi-lingual tokenizers have to share the capacity across multiple languages.

To evaluate the impact, we trained the mono-lingual Ukrainian models with two different tokenizers, see Table~\ref{tab:tokenizers_comparison}.
We conclude that use of a mono-lingual tokenizer indeed provides better accuracy and we consider the WERR of 18.1\% as our baseline.

\subsection{Cross-lingual training}

\subsubsection{Knowledge transfer from English}

Cross-lingual knowledge transfer to unseen languages has been shown to be successful, e.g., in~\cite{cho2018multilingual}. We hence trained two bi-lingual Ukrainian and Russian models using the balanced dataset, one trained from scratch and the other with the encoder initialized from a strong Transducer-based English model. Since the English model uses a different tokenizer, we re-initialized the label encoder and the joiner network. Based on the results in Table~\ref{tab:knowledge_transfer_from_English}, we conclude that use of the English prior model has negative impact on the accuracy, achieving 18.1\% WERR compared to 20.8\% achieved by the model trained from scratch.

\begin{table}[th]
  \caption{Effect of knowledge transfer from a strong English prior model}
  \label{tab:knowledge_transfer_from_English}
  \centering
  \begin{tabular}{ l r r }
    \toprule
Warm-start & - & English  \\
Data used & \multicolumn{2}{c}{UKR + RUS BL} \\
Training data & &  \\
UKR / Total [khrs] & 	\multicolumn{2}{c}{25 / 41} \\
Tokenizer & \multicolumn{2}{c}{2 Langs} \\
    \midrule
Model ref & 2L-BL & E-2L-BL  \\
    \midrule
 WERR [\%] & \textbf{20.8} & 18.1  \\
    \bottomrule
  \end{tabular}
\end{table}

\subsubsection{Knowledge transfer from Slavic languages}

 \begin{table*}[th]
  \caption{Comparison of mono-lingual Ukrainian model with multi-lingual models trained with naturally weighted (NW) and balanced (BL) datasets.}
  \label{tab:crosslingual_semisup}
  \centering
  \begin{tabular}{ l r r r r r }
    \toprule
Data used	& 5 Langs NW &	5 Langs BL & UKR + RUS NW & UKR + RUS BL & UKR \\
Training data & & & & & \\
UKR / Total [khrs] & 	12 / 636 &	25 / 109 &	12 / 513 &	25 / 41 & 25 / 25 \\
Tokenizer &	5 Langs & 5 Langs & 2 Langs & 2 Langs  & UKR \\
    \midrule
Model ref & 5L-NW & 5L-BL  & 2L-NW & 2L-BL & U-BL \\
    \midrule
\multicolumn{6}{c}{Language specific WERR [\%]} \\
    \midrule
UKR & 0.0 & 19.4 & 0.7 & \textbf{20.8} & 18.1 \\
RUS & 16.5& -12.4 & 16.5 & -8.2	 & - \\
POL & 9.1 & 11.7 & - & -  &  - \\
CZE & -2.9 & 5.9 & - & -  &  - \\
SVK & -4.2 & 11.5 & - & -  &  - \\
    \bottomrule
  \end{tabular}
\end{table*}

Next, we investigate the effect of cross-lingual knowledge transfer from the related Slavic languages. For the multi-lingual models, we compare the effect of training with the naturally weighted and balanced datasets. The former provides more data but predominately Russian, which is not our target language.

In this, and all the subsequent experiments, regardless of the amount of unique training data, the training is scheduled for a specific number of training steps. We evaluate models from multiple steps of the training and pick the step that provides best accuracy. This is done to account for a possible difference in convergence speed with respect to different amount of unique training data and possible fluctuation of accuracy across languages during the training. 

We evaluate WERR for all five languages included in the training data with respect to their reference established by monolingual DNN-HMM hybrid systems, see Table~\ref{tab:crosslingual_semisup}.
We conclude that for all languages, except for Russian, training with the balanced datasets provides clearly better results.
For Russian, on the other hand, training with the naturally weighted data produces a model with notably better accuracy (WERR of 16.5\% vs. -12.4\%), thanks to much more data available for this language.

Focusing on WERR for Ukrainian, first, we see that both multi-lingual models trained using the balanced datasets outperform the mono-lingual model (18.1\% WERR).
The bi-lingual model trained using Russian and Ukrainian data (20.8\% WERR) outperforms the model trained using all languages (19.4\% WERR), which suggests that the language proximity of the two languages is beneficial.

\subsubsection{Fine-tuning to Ukrainian}

Our goal is not to build multi-lingual models but primarily use cross-lingual knowledge transfer for improving accuracy in the target language. Hence, we fine-tuned the multi-lingual models with Ukrainian only data, see Table~\ref{tab:crosslingual_semisup_finetune} for results.
Interestingly, after the fine-tuning, ranking of the models from Table~\ref{tab:crosslingual_semisup}, notably changes.

\begin{table*}[th]
  \caption{Comparison of multi-lingual models fine-tuned with Ukrainian data only}
  \label{tab:crosslingual_semisup_finetune}
  \centering
  \begin{tabular}{ l r r r r }
    \toprule
Warm-start & 5L-NW & 5L-BL & 2L-NW & 2L-BL \\
Data used & \multicolumn{4}{c}{UKR (25khrs)} \\
Tokenizer & 5 Langs & 5 Langs & 2 Langs & 2 Langs \\
    \midrule
Model ref & 5L-NW-U & 5L-BL-U  & 2L-NW-U & 2L-BL-U \\
    \midrule
 WERR [\%] & \textbf{25.0} & 21.5 & 24.3 & 	22.2 \\
    \bottomrule
  \end{tabular}
\end{table*}

We find that using the multi-lingual models trained using the naturally weighted data to initialize the fine-tuning leads to better results.
Also, the advantage of bi-lingual models over the models trained with all languages is not as clear anymore.

In fact, the best accuracy is achieved by the multi-lingual model trained with the naturally weighted data from all languages and fine-tuned to Ukrainian, suggesting that the model's accuracy benefits from availability of a larger volume of data from more diverse, though still related, set of languages.
The fine-tuning results also demonstrate the impact of learning curriculum along with the importance of availability of the data.
Overall, we were able to achieve WERR of 25\% (`5L-NW-U`) using the cross-lingual knowledge transfer compared to 18.1\% (`U-BL`) achieved in mono-lingual training.

\subsection{Iterative Pseudo-labeling}

Observing the 25\% WERR compared to the hybrid DNN-HMM system, we re-decoded the Ukrainian data with the new model and we initialized training with the model from the previous stage.

\subsubsection{First pass re-decoding}

Besides the improved accuracy, there is another advantage of using the new models for re-decoding.
Unlike the original hybrid model, our new Transducer models provide log probability for each output token, which can be used as a certainty metric.
We explored different ways to combine the token-level log probabilities into an query level metric and found summation to work best.
Table~\ref{tab:crosslingual_semisup_finetune_1st_redecode} summarizes the results for two models trained using the re-decoded Ukrainian data.
We compare training using all data and training using 50\% of the queries with highest certainty.
We see that the model trained using the data selected based on the certainty criteria (32.6\% WERR) outperforms that model trained using all the data (29.9\% WERR), despite using only half of the data.

\begin{table}[th]
  \caption{Results after first round of re-decoding with Transducer model, with and without certainty selection criteria applied}
  \label{tab:crosslingual_semisup_finetune_1st_redecode}
  \centering
  \begin{tabular}{ l r r }
    \toprule
Warm-start & \multicolumn{2}{c}{5L-NW-U} \\
Data used & \multicolumn{2}{c}{1st pass re-decoded UKR} \\
Certainty range & N/A & 50-100\%  \\
Training data [khrs] & 25 & 12.5 \\
    \midrule
Model ref & 5L-NW-U-P1A & 5L-NW-U-P1C   \\
    \midrule
 WERR [\%] & 29.9 &	\textbf{32.6} \\
    \bottomrule
  \end{tabular}
\end{table}

\begin{table}[th]
  \caption{Results after second round of re-decoding with Transducer model, with and without certainty selection criteria applied}
  \label{tab:crosslingual_semisup_finetune_2nd_redecode}
  \centering
  \begin{tabular}{ l r r }
    \toprule
Warm-start &  \multicolumn{2}{c}{5L-NW-U-P1C} \\
Data used & \multicolumn{2}{c}{2nd pass re-decoded UKR} \\
Certainty range & N/A & 50-100\%  \\
Training data [khrs] & 25 & 12.5 \\
    \midrule
Model ref & 5L-NW-U-P2A & 5L-NW-U-P2C   \\
    \midrule
 WERR [\%] & \textbf{34.7} & \textbf{34.7} \\
    \bottomrule
  \end{tabular}
\end{table}

\subsubsection{Second pass re-decoding}

Since the model fine-tuned using the re-decoded data notably outperforms the model from the earlier stage (`5L-NW-U`), we re-decoded the data one more time.
Finally, we built models using the transcripts from the second re-decoding pass and using the model from the previous stage (`5L-NW-U-P1C`) to initialize the training.

We again trained models using all the data and then 50\% of the queries for which the model was most certain.
This time, we do not see any impact of the certainty based data selection.
One possible explanation is that the way the token level certainty is combined into an query level metric is not effective in the second re-decoding stage as we observed the distribution of the log-probabilities produced by the models to shift into a much narrower range between the first and second re-decoding stage. 

Comparing the accuracy gain achieved by the first and second re-decoding passes, we see that the effect diminishes and hence we did not carry out more re-decoding iterations.
However, overall, we were able to achieve 34.7\% WERR compared to the reference established by the hybrid system.

\section{Conclusions}

First, we can built a Transducer-based system achieving 18.1\% relative WER reduction compared to a reference DNN-HMM hybrid system. To train the Transducer-based system, we used only semi-supervised data with automatic transcripts produced by the hybrid system and no manually annotated data.

More importantly, we have demonstrated that combination of the cross-lingual knowledge transfer and iterative pseudo-labeling can further improve the accuracy, ultimately achieving 34.7\% relative improvement over the hybrid reference system.

One aspect worth highlighting is that in the multi-stage training strategy, the models that underperform in one stage, may still provide a strong prior model for the subsequent fine-tuning stage.
In particular, our results show that model trained on large volume of data from non-target languages can provide better initialization for fine-tuning with the data from the target language, even though the model itself underperformed competing candidates.
This also highlights the importance of the training curriculum applied along with the data used.

Despite the proximity of the languages in our investigation, the languages differ significantly in terms of the baseline accuracy and amount of available data.
In the future, we would like to expand the investigation on all the languages.
We are also planning to investigate the effect of application of the cross-lingual knowledge transfer in iterative fashion.

\section{Acknowledgements}

We would like to thank Tatiana Likhomanenko, Pawel Swietojanski, Ossama Abdelhamid and Barry Theobald for valuable discussions.

\bibliographystyle{IEEEtran}
\bibliography{mybib}

\end{document}